\documentclass[runningheads]{llncs}
\usepackage[T1]{fontenc}
\usepackage{graphicx}
\usepackage[utf8]{inputenc}
\usepackage[english]{babel}

\usepackage[numbers]{natbib}
\usepackage{multirow}
\usepackage{amssymb}
\usepackage{diagbox}
\usepackage{enumitem}
\usepackage{amsmath}
\usepackage{mathrsfs}  
\usepackage{amsfonts}
\usepackage{commath}
\usepackage{balance}
\usepackage{xspace}
\usepackage[noend]{algpseudocode}
\usepackage[linesnumbered,ruled,vlined, noend]{algorithm2e}
\usepackage{tabularx}
\usepackage{graphicx,subfigure}

\usepackage{inconsolata}

\usepackage{graphicx}

\usepackage{tikz}
\usepackage{xspace}
\usepackage{booktabs}
\usepackage{pifont}
\usepackage{multirow}
\usepackage{csquotes}
\usepackage{anyfontsize}
\usepackage{subcaption}
\usepackage{enumitem}
\usepackage{tabularx}
\usepackage{makecell}

\usepackage{adjustbox}
\usepackage{breqn}
\newcommand{\stitle}[1]{\noindent\textbf{#1.}}

\definecolor{Lightgray}{RGB}{110,110,110}

\begin{document}
\title{Counterspeech for Mitigating the Influence of Media Bias: Comparing Human and LLM-Generated Responses}
\titlerunning{Counterspeech for Mitigating the Influence of Media Bias}
\author{Luyang Lin\inst{1,2} \and
Zijin Feng\inst{1} \and
Lingzhi Wang\inst{3} \and
Kam-Fai Wong\inst{1,2}}
\authorrunning{L. Lin et al.}
\institute{The Chinese University of Hong Kong, Hong Kong, China \\
% \url{http://www.springer.com/gp/computer-science/lncs}
\and
MoE Key Laboratory of High Confidence Software Technologies, China \\
% \email{\{lylin, zjfeng, kfwong\}@se.cuhk.edu.hk}\\
% \url{http://www.springer.com/gp/computer-science/lncs}
\and
Harbin Institute of Technology, Shenzhen, China\\
\email{\{lylin, zjfeng, kfwong\}@se.cuhk.edu.hk}  \qquad 
\email{wanglingzhi@hit.edu.cn }}
\maketitle              % typeset the header of the contribution

\textit{\textbf{Warning}: This paper contains content that may be offensive or controversial.}
\vspace{-0.8cm}
\begin{abstract}
Biased news contributes to societal polarization and is often reinforced by hostile reader comments, constituting a vital yet often overlooked aspect of news dissemination. Our study reveals that offensive comments support biased content, amplifying bias and causing harm to targeted groups or individuals.
Counterspeech is an effective approach to counter such harmful speech without violating freedom of speech, helping to limit the spread of bias. To the best of our knowledge, this is the first study to explore counterspeech generation in the context of news articles. We introduce a manually annotated dataset linking media bias, offensive comments, and counterspeech. 
We conduct a detailed analysis showing that over 70\% offensive comments support biased articles, amplifying bias and thus highlighting the importance of counterspeech generation.
Comparing counterspeech generated by humans and large language models, we find model-generated responses are more polite but lack the novelty and diversity. Finally, we improve generated counterspeech through few-shot learning and integration of news background information, enhancing both diversity and relevance~\footnote{The dataset will be released when the paper is published.}.

\end{abstract}
\section{Introduction}

Although the theory of objectivity is considered a cornerstone principle of journalism, its achievability has been contested~\cite{ryan2001journalistic, munoz2012truth}.
As a result, biased news continues to be produced and disseminated via social media, fueling societal polarization~\cite{bernhardt2008political} and affecting key societal decisions~\cite{watts2021measuring, bernhardt2008political, druckman2005impact}.
Notably, news does not spread in isolation: reader interaction, particularly via comments, accompanies the article and influences how other readers perceive and react to it~\cite{stroud2016news, spinde2023twitter}. This dynamic contributes to the prevalence of hostile emotions in comments within polarized information environments~\cite{humprecht2020hostile,spinde2023twitter}, suggesting that understanding media bias requires considering both the article 
% content 
and comments.

Offensive comments intensify the harm of biased content for targeted individuals or groups and may escalate to hate crimes~\cite{olteanu2018effect}. This issue has received increasing attention, with many studies focusing on the automatic detection of hate speech~\cite{vargas2021hatebr,vidgen2021introducing}. However, despite advances in detection, efforts by social media platforms to combat hate speech through blocking or suspending offending messages or accounts have proven to have limited effectiveness. Counterspeech has thus emerged as a more promising approach for addressing harmful speech without undermining the freedom of speech~\cite{richards2000counterspeech, mathew2019thou}.

Addressing media bias and offensive language in isolation is inadequate for improving the information environment, as harmful discourse often intensifies through reader comments. Therefore, it is crucial to extend the focus to counterspeech generation to mitigate the influence of bias. Most prior research on media bias in literature focuses on the biased content itself or its relation to news outlets~\cite{baly2020we} and contextual factors~\cite{van2020context, lin2023data}. Only a few studies explore the relationship between article contents and comments to analyze the hate speech~\cite{spinde2023twitter}. To our knowledge, no prior work has leveraged media bias, offensive comment, and counterspeech simultaneously to mitigate the impact of media bias.

%%%%%%%%%%%%%%%%%%%%%%%%%%%%%%%
\begin{table}[t]
\begin{center}
\small
\begin{tabular}{lp{8.5cm}}
\toprule
% \textbf{Strategy} & \textbf{Human} & \textbf{LLM}\\
% \midrule
\textbf{News} & Rep. Gaetz Protesting Canceled California Rallies: 'We Will Not Be Silenced' \\\midrule
\textbf{Bias Label} &  Skews Right \\\midrule
\textbf{Reliability} &  Mixed Reliability \\\midrule

\textbf{Comment} & Don’t you guys think every Democrat is a sex criminal despite no evidence and no one in jail? \\\midrule
\textbf{Relation} &  Support \\\midrule
\textbf{Reply} & I don't know who the you guys are that you're speaking of but I tried to go on actual evidence. My point being if there is real evidence why is no one in jail?\\ \bottomrule

\end{tabular}
\end{center}
\vspace{-0.2cm}
\caption{Illustrative example of a news article annotated with bias and reliability, an offensive comment supporting the article's stance, and a counterspeech reply using the \textit{Counter Question} strategy.}
\vspace{-1.2cm}
\label{tab:intro_case}
\end{table}

%%%%%%%%%%%%%%%%%%%%%%%%%%%%%%%

To fill this gap, we propose an annotation framework to label offensive comments and their corresponding counterspeech. In the case shown in Table~\ref{tab:intro_case}, under our framework, the comment is annotated as supporting the article’s stance, showing offensiveness to Democrats, and the reply is labeled as counterspeech using a \textit{Counter Question} strategy highlighting the lack of evidence. Our dataset comprises $853$ offensive comment - counterspeech reply pairs from $279$ news articles across $93$ outlets.
Our analysis reveals that most offensive comments support the original article, underscoring the critical role of comments in studying offensive language. Additionally, $16\%$ of human counterspeech replies employ Hostile Language, which may not improve the conversation environment~\cite{howard2021terror}.

We further contribute a benchmark for counterspeech generation by conducting experiments using Large Language Models (LLMs) on our dataset. Compared with the human-written counterspeech, the proportion of the Positive Tone strategy has increased significantly from $12.9\%$ to $20.2\%$. In human evaluation, $77.4\%$ of annotators found LLM-generated replies more persuasive when they were asked to put themselves in the role of senders of offensive comments. However, the LLM-generated replies generally lack diversity. To address this, we explore techniques such as few-shot prompting with diverse examples, specifying a strategy, or including article information. Main contributions of this paper are:
\vspace{-0.5cm}
\begin{itemize}[itemsep=0.1cm,parsep=0.0cm]
    \item We introduce a manually annotated dataset capturing the relationship among media bias, offensive comments, and counterspeech replies, filling a key gap and being the first to leverage them for counterspeech generation.
    \item We conduct a detailed analysis showing that 70\% offensive comments support biased articles, amplifying bias and thus highlighting the importance of counterspeech generation. Compared to human replies, LLM-generated counterspeech is notably more polite--with a 57\% more Positive Tone and a 57\% less Hostile Language--but lacks diversity by 69\%.

    \item We establish a counterspeech generation benchmark on media bias, and investigate techniques that improve LLM-generated counterspeech novelty and diversity by 17.6\%.
\end{itemize}
\vspace{-0.5cm}

\section{Related Work}

% \vspace{-0.3cm}
\subsection{Media Bias}
% \vspace{-0.2cm}
Media bias can be defined as ``imbalance or inequality of coverage rather than as a departure from truth'' \cite{stevenson1973untwisting} and ``any systematic slant favoring one candidate or ideology over another'' \cite{waldman1998newspaper} at the beginning. Although there is no universal definition for media bias, now it is frequently defined as slanted news coverage or internal bias, reflected in news articles \cite{hamborg2019automated}.

Facing this problem, most of the previous work focuses on the biased text itself to detect the media bias \cite{baly2020we, lim2020annotating, spinde2022neural} from lexical and informational perspectives \cite{fan2019plain,lin2023data,lin2024indivec}, or explores flipping the bias \cite{chen2018learning}.
Some analyze the media bias in LLMs and debiasing methods \cite{feng2023pretraining,lin2024investigating,fulay2024relationship}.
Few take the reader into account \cite{spinde2023twitter}, although comments are also an important part of news dissemination. Our work not only considers the comments but also explores the generation of counterspeech to deduce the influence of harmful comments, which completes the entire media bias research system. 

\vspace{-0.3cm}
\subsection{Offensive Language and Counterspeech}
Previous research shows that biased news will result in hostile comments \cite{humprecht2020hostile,spinde2023twitter}, and the effective way to solve this problem is counterspeech \cite{richards2000counterspeech, mathew2019thou}, also, it can be used to counter misinformation \cite{micallef2020role, saha2024integrating}. 
However, previous research about offensive language and its counterspeech is isolated from the news article \cite{zhu2021generate,saha2024crowdcounter,saha2024zero}, which cannot present the causation target hate speech during the spread of biased articles. To the best of our knowledge, there has been no prior work that explores counterspeech under biased news, and our contributed dataset can fill this gap. We also apply different strategies \cite{benesch2016counterspeech,mathew2019thou,saha2024integrating} to analyze the counterspeech in a fine-grained way.

% \vspace{-0.3cm}
% \subsection{LLM Generation}
% Most of the previous work focuses on the detection of hate speech and counterspeech \cite{wich2020impact, yu2022hate}.
% Due to the strong ability of LLM, more researchers use LLM to explore the generation task \cite{zhu2021generate, bonaldi2023weigh, saha2024integrating}. Some of them applied zero-shot or few-shot prompting \cite{saha2024zero, ashida2022towards}.
% In this work, we add more news background to explore its effect on counterspeech generation.
\section{Dataset}
\vspace{-0.3cm}
To better analyze the relations among media bias, offensive comments, and counterspeech, we make an extension based on Bias And Twitter dataset (BAT) \cite{spinde2023twitter}.
\vspace{-0.9cm}
\subsection{BAT Overview}
\vspace{-0.2cm}
This dataset contains both news articles and their comments.
The articles are from Ad Fontes Media’s website\footnote{https://adfontesmedia.com/rankings-by-individual-news-source/; accessed on 2021-10-26} and have respective scores of political bias and reliability based on each article, which have been suggested as high-quality labels \cite{chen2020analyzing}.
Specifically, the articles are divided into 4 reliability classes (Reliable, Generally Reliable, Mixed Reliability, and Unreliable) and 5 bias classes (Hyper-Partisan Left, Skews Left, Middle, Skews Right, and Hyper-Partisan Right).

The comments on news articles are from Twitter, and outlets’ tweets referencing the rated articles, together with quoted retweets, are collected.
In total, there are $2,800$ news articles from $255$ different English-speaking news outlets and $175,807$ comments and retweets referring to these articles.
\vspace{-0.5cm}
\subsection{Counterspeech Strategy}
\label{subsec:counterspeech_strategy}
\vspace{-0.2cm}
We define the counterspeech following \citep{yu2022hate} as the author challenges, or condemns the hate expressed in another comment, or calls out a comment for being hateful.

The definitions of response strategy are first proposed by \citep{benesch2016counterspeech}, and we follow \citep{mathew2019thou,saha2024integrating} to refine them into 9 categories: \textit{1. Affiliation}: Depicting a positive affective relationship; \textit{2. Counter question:} Questioning the message; \textit{3. Denouncing:} Denounce the message as being hateful; \textit{4. Hostile language:} Abusive, hostile, or obscene response; \textit{5. Humor and sarcasm:} Humorous or sarcastic response; \textit{6. Pointing out hypocrisy or contradiction:} Points out the hypocrisy or contradiction; \textit{7. Positive tone:} Use empathic, kind, polite, or civil speech; \textit{8. Presenting facts:} Tries to persuade by correcting misstatements \textit{9. Warning of consequences:} Warns of possible consequences of actions.

\vspace{-0.5cm}
\subsection{Structure Design}
\vspace{-0.8cm}
\begin{figure*}
    \centering
    \includegraphics[width=0.95\linewidth]{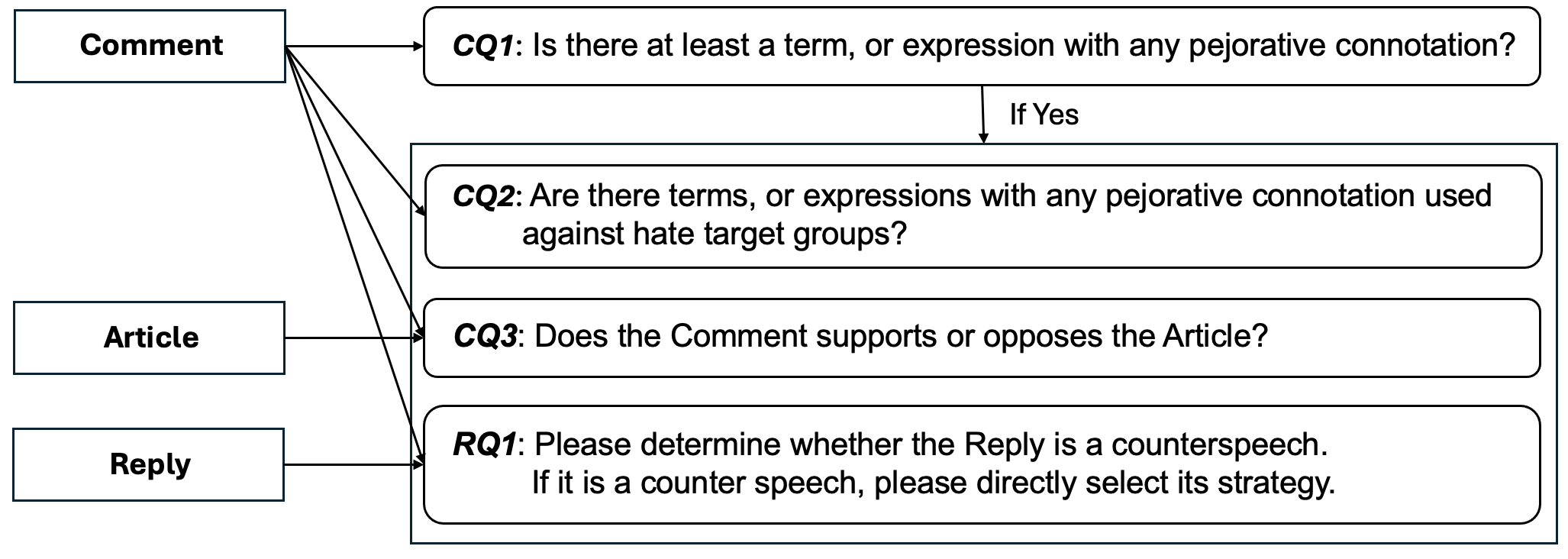}
    \vspace{-0.3cm}
    \caption{Design of our annotation structure}
    \vspace{-0.3cm}
    \label{fig:annotation_structure}
\vspace{-0.5cm}
\end{figure*}
\vspace{-0.1cm}
We designed the annotation structure for our task as shown in Figure~\ref{fig:annotation_structure}. It contains two main modules, one is offensive language and hate speech annotation, with the relation between the hate speech and the news article, and another is counterspeech and strategy annotation. 

In the hate speech annotation, we follow \citep{vargas2021hatebr} to split it into two questions, \textit{CQ1} is to label whether it is an offensive language by checking pejorative connotation terms or expression and \textit{CQ2} is to confirm whether it is a hate speech by checking whether the offensive language is targeting group. Then we will supply the headline and URL of the original news article, to get the relation between the comment and the original news, i.e., supports or opposes (\textit{CQ3}).  
In the counterspeech annotation module, the reply of this comment will be given to annotate whether it is a counterspeech, and if yes, the annotator will be asked to classify the strategy the counterspeech used (\textit{RQ1}).

We divided the whole process into two stages. In the first stage, \textit{CQ1} is asked, and only if the answer is yes, which means that it contains offensive language, the reply will go into the second stage to be labeled by the other questions.
\vspace{-0.5cm}
\subsection{Annotation}

\vspace{-0.2cm}
\subsubsection{Dataset Preprocessing}
We preprocess the dataset BAT to make it better fit our task. First, we rearrange the comments and only keep those with at least one reply. Then we remove the non-text characters, like emoji, and the user information in the text is also removed. Finally, 21,450 comments are left

\vspace{-0.6cm}
\subsubsection{Preliminary Annotation}
Due to the high cost of human annotation, we use LLMs to do the preliminary annotation. We prompt \textit{gpt-3.5-turbo} and \textit{gpt-4o-mini} by zero-shot to answer the \textit{CQ1}. $1,258$ comments are labeled as offensive language by both two models and will be further confirmed by human annotators.

\vspace{-0.6cm}
\subsubsection{Human Annotation}
\label{subsec:huamn_annotation}
We have two in-house annotators to conduct the annotation process and employ the annotators from Amazon Mechanical Turk (MTurk) \footnote{https://www.mturk.com/}. We only allow Mechanical Turk Masters\footnote{A Master Worker is a top Worker of the MTurk marketplace that has been granted the Mechanical Turk Masters Qualification} having a high approval rate ($95\%$) to participate. In this task, each question will be annotated by at least 2 annotators. After the annotation process, our in-house annotators will help to ensure the quality of the dataset and make the final decisions.

We use Cohen's Kappa to measure the inter-annotator agreement (IAA) \cite{cohen1960coefficient, artstein2008inter}. In the first stage, the calculated Kappa value for \textit{CQ1} is $0.605$, and $891$ comments are labeled offensive language for further processing.
In the second stage, we have an overall value of $0.517$ for three questions, $0.536$, $0.421$ for \textit{CQ2}, \textit{CQ3} separately.
In question \textit{RQ1}, Cohen's Kappa for binary classification of counter and non-counterspeech is $0.583$, and in fine-grained classification, it is $0.352$ for different strategies selections. Although it is a complex task to distinguish the strategies by crowdsourcing annotators, we also reach a fair agreement.

\vspace{-0.5cm}
\subsection{Statistics}
\vspace{-0.2cm}
\label{subsec:dataset_statistics}
Our final dataset consists of $853$ pairs, each comprising a Twitter comment with offensive language and its corresponding counterspeech reply. These pairs are under $279$ news articles covering 93 different media outlets. Notably, $93.7\%$ of the offensive comments are regarded as hate speech.

\paragraph{\textbf{Offensive Language}}

\begin{figure}[t]
\vspace{-0.2cm}
  \centering
  \subfigure[Overview]{
      \includegraphics[width=0.3\linewidth]{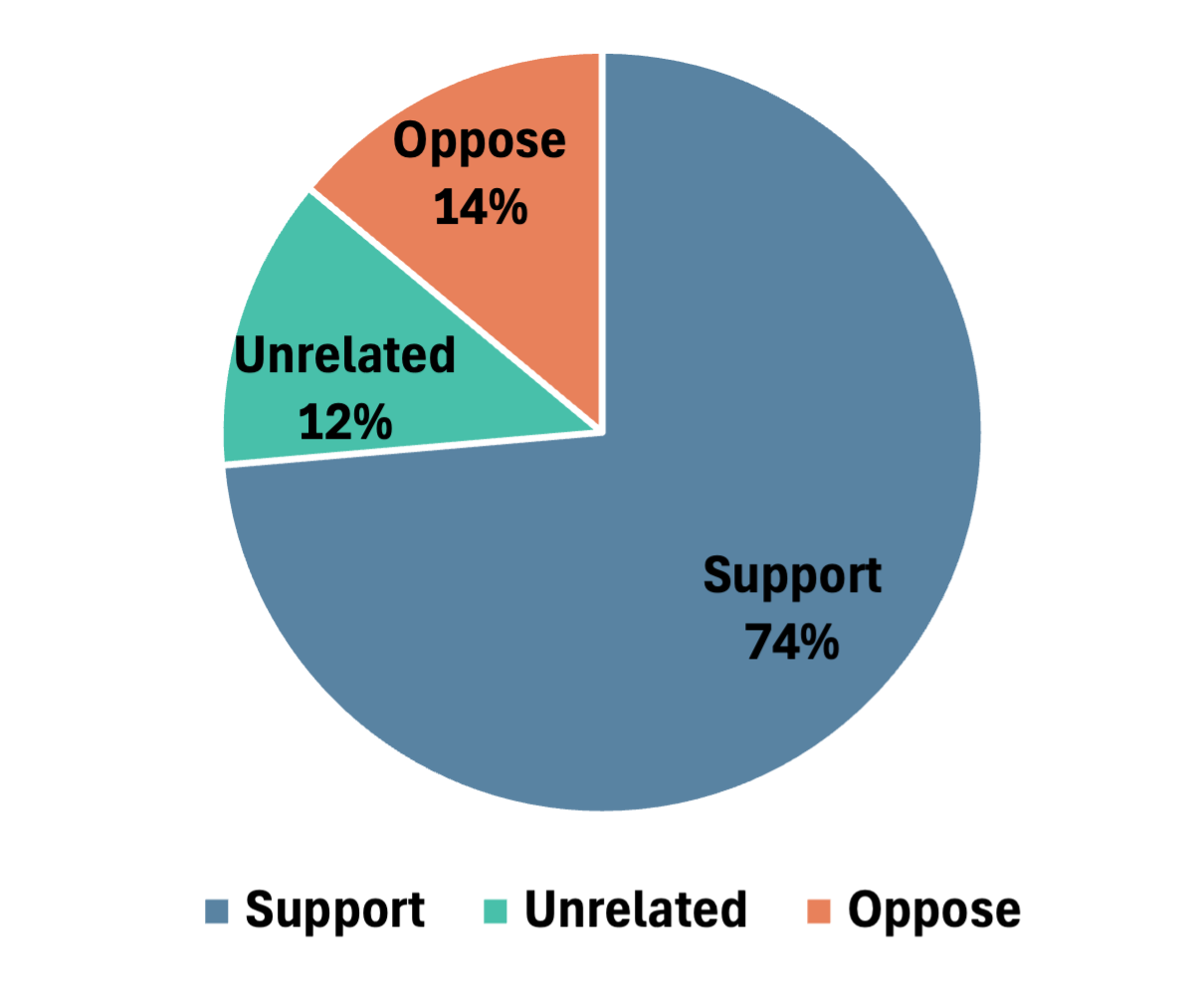}
      % \vspace{-0.1cm}
      \label{fig:stat_offensive1}
  }
  % \vspace{-0.3cm}
  \subfigure[Reliability]{
      \includegraphics[width=0.3\linewidth]{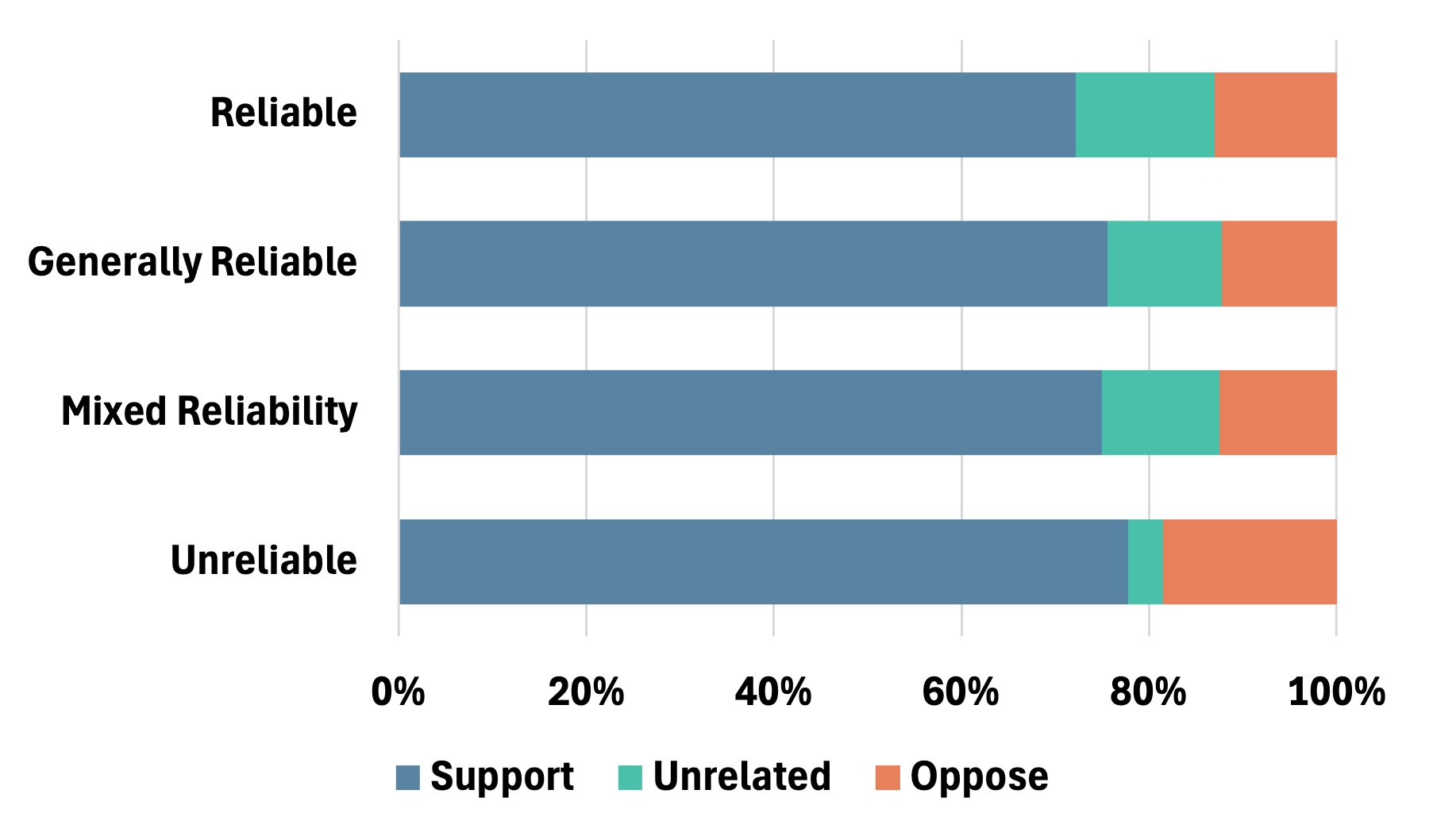}
      % \vspace{-0.1cm}
      \label{fig:stat_offensive2}
      
  }
  \subfigure[Bias]{
      \includegraphics[width=0.3\linewidth]{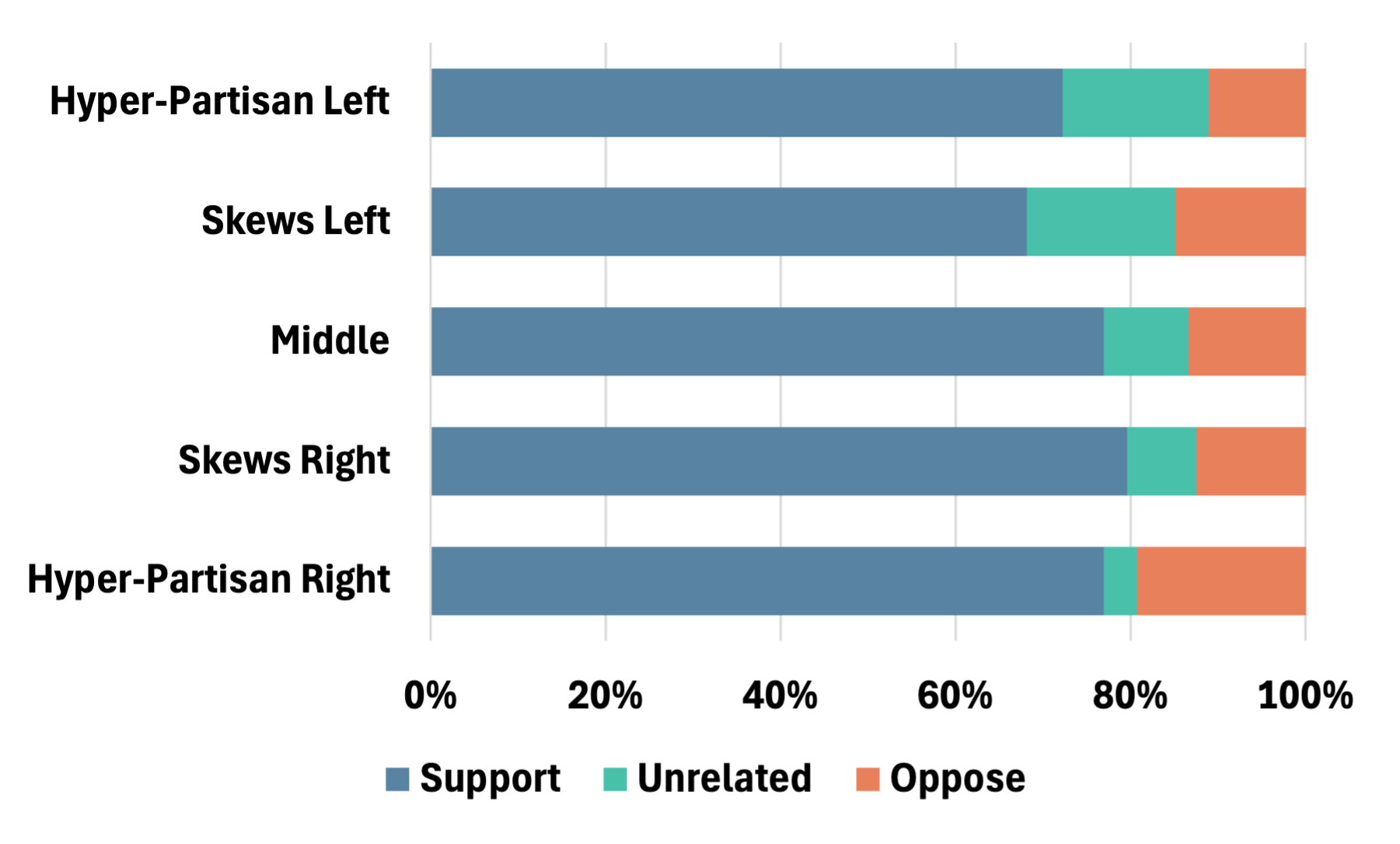}
      % \vspace{-0.1cm}
      \label{fig:stat_offensive3}
  }

\vspace{-0.3cm}
\caption{\label{fig:stat_offensive}Statistics in the relationship between the offensive comment and the original news article. The classifications of the articles in reliability and bias are from Ad Fontes Media.}
\vspace{-0.8cm}
\end{figure}

We analyze the relationship between offensive comments and original news articles by whether the comment supports or opposes the content, and the results are shown in Figure~\ref{fig:stat_offensive}.
From the overview in Figure~\ref{fig:stat_offensive1}, we observe that most of the comments support the original articles, which indicates that to some extent, readers are venting their emotions along with the news content. Also, there are $12\%$ comments that are not related to original articles.

We also notice that whether the article is reliable or not, or middle or not, there are quite a few hateful comments. Because the original dataset BAT does not have balanced data in the distribution of bias and reliability, with more left-leaning and reliable news \cite{spinde2023twitter}, we use the percentage instead of the number of comments to present the relationship in Figure~\ref{fig:stat_offensive2} and Figure~\ref{fig:stat_offensive3}.

We can observe that if the news articles are unreliable, more readers will leave comments to oppose them, and the proportion is nearly twice that of the comments in reliable articles. As for the comments in articles of different biased classes, more hate comments will support the news content in right-leaning rather than left-leaning articles.

\begin{figure}
\centering
\vspace{-0.8cm}
\includegraphics[width=0.6\linewidth]{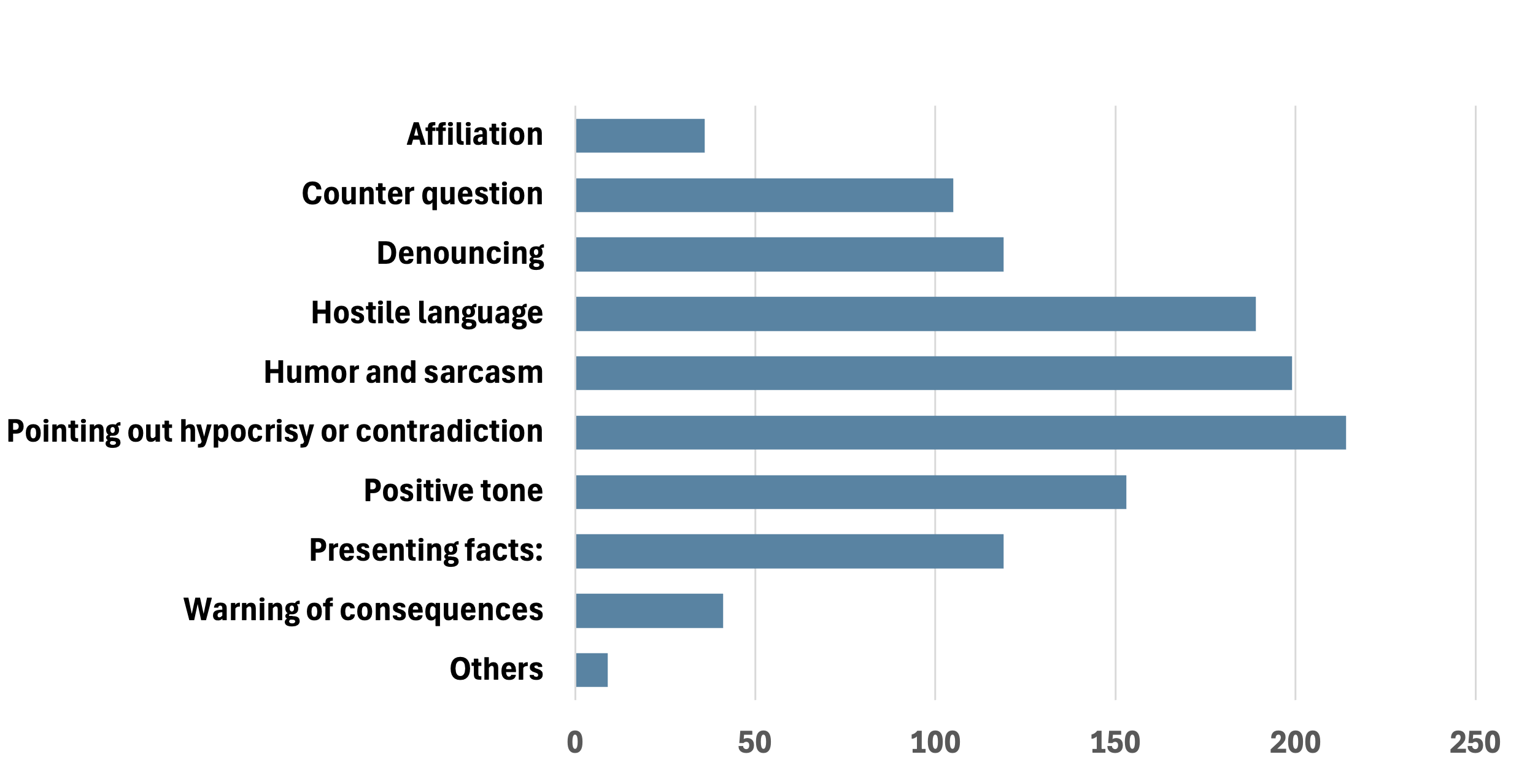}
\vspace{-0.3cm}
\caption{\label{fig:stat_counter}Statistics in various counterspeech strategies}
\vspace{-0.8cm}
\end{figure}

In conclusion, polarized and harmful opinions may be caused by information cocooning \cite{jiang2025information}, so it is more valuable to break the informational barriers and counter hate speech.

\vspace{-0.4cm}
\paragraph{\textbf{Counterspeech}}

In the replies to the offensive comments, $95.7\%$ can be regarded as counterspeech, and the distribution of the strategy is shown in Figure~\ref{fig:stat_counter}. The strategy that is used most is \textit{Pointing out hypocrisy or contradiction}. We can also observe that \textit{Hostile language} is used more than \textit{Positive tone}. Although the harm has been pointed out, the conversation environment still cannot become as friendly as we expect.
\vspace{-0.4cm}

\section{Methodology}

\vspace{-0.2cm}
\subsection{Task Design}
\vspace{-0.1cm}
We design several tasks to test the ability of LLMs in generating counterspeech.
\vspace{-0.7cm}

\paragraph{\textbf{Vanilla Counterspeech Generation}}
In this task, we directly give the LLM a comment and prompt it to generate a counterspeech without other instructions.  
It is designed to compare the differences in the counterspeech strategies used between the human-written and model-generated replies.

\vspace{-0.3cm}
\paragraph{\textbf{Strategy-Based Generation}}
In this task, we will specify a type of strategies for the LLM to generate the counterspeech, and the details of the descriptions of the strategies are in Section~\ref{subsec:counterspeech_strategy}. Our experiments will be conducted on both zero-shot and few-shot settings. The examples for the few-shot are randomly selected, and the prompt is structured as follows:

\vspace{-0.2cm}
\begin{quote}
\setlength\leftskip{-0.4cm}
\textit{\small
Counterspeech is a strategic response to hate speech, aiming to foster understanding or discourage harmful behavior. You are an instruction-following, helpful AI assistant that generates counter-hate speech replies. Help as much as you can. You will first be given \colorbox{Lightgray}{\color{white}{\textit{N}}} examples, each example consists of a hate comment, a counter-hate speech strategy, and a counter-hate speech reply. Then you will be given a hate comment and a counter-hate speech strategy, Please write a counter-hate speech reply following the style of the examples.
\colorbox{Lightgray}{\color{white}{\textit{EXAMPLES}}}\\
Given the following hate comment and strategy, please write a counter-hate speech reply following the style of the above examples. Write me the reply only.\\
Hate comment: \colorbox{Lightgray}{\color{white}{\textit{COMMENT}}}
Counter hate speech strategy: \colorbox{Lightgray}{\color{white}{\textit{STRATEGY}}}
}
\end{quote}

\vspace{-0.5cm}
\paragraph{\textbf{News Background Consideration}}
In this task setting, the news background of the offensive comment will be considered, because we mentioned that most of the hate comments are closely related to the original articles Section~\ref{subsec:dataset_statistics}. The title of the news article will be given in this setting, and the experiments will be conducted on zero-shot and few-shot together with strategy-based and non-strategy-based settings. The prompt is structured as follows:

\vspace{-0.2cm}
\begin{quote}
\setlength\leftskip{-0.4cm}
\textit{\small
Counterspeech is a strategic response to hate speech, aiming to foster understanding or discourage harmful behavior. You are an instruction-following, helpful AI assistant that generates counter-hate speech replies. Help as much as you can. Below, you will first be given \colorbox{Lightgray}{\color{white}{\textit{N}}} examples, each example consists of 1. a hate comment, 2. the title of the original news article where the comment was posted, 3. a counter hate speech strategy, and 4. a counter hate speech reply. Then you will be given 1. a hate comment, 2. the title of the original news article where the comment was posted, and 3. a counter-hate speech strategy. Please write a counter-hate speech reply following the style of the examples. 
\colorbox{Lightgray}{\color{white}{\textit{EXAMPLES}}}\\
Given the following hate comment and strategy, please write a counter-hate speech reply following the style of the above examples. Write me the reply only.\\
Hate comment: \colorbox{Lightgray}{\color{white}{\textit{COMMENT}}}
Title of the original news article: \colorbox{Lightgray}{\color{white}{\textit{TITLE}}}\\
Counter hate speech strategy: \colorbox{Lightgray}{\color{white}{\textit{STRATEGY}}}
}
\end{quote}

\vspace{-0.7cm}

\subsection{Model}

\vspace{-0.2cm}
\begin{table*}[t]
\vspace{-0.3cm}
\begin{center}
\scalebox{0.88}{
\begin{tabular}{l|c|c|c|c|c|c|c|c}
\toprule
% \multirow{2}{*}{Type}
\textbf{Dataset} & \textbf{size} & \textbf{length} & \textbf{fk $\downarrow$} & \textbf{dc $\downarrow$} & \textbf{div $\uparrow$} & \textbf{arg $\uparrow$} & \textbf{c-arg $\uparrow$} & \textbf{tox $\downarrow$} \\
 
\midrule	

Gab \cite{qian2019benchmark} & 13678 & 15.54 & 8.67 & \textbf{\underline{8.55}} & 0.73 & \textbf{\underline{0.17}} & 0.47 &  0.15\\

Reddit \cite{qian2019benchmark} & 5203 & 16.03 & 8.80 & 8.70 & 0.72 & \textbf{\underline{0.17}} & 0.44 &  0.14\\

% CrowdCounter & 1325 & 20.65 & 7.64 & 8.35 & 0.85 & 0.21 & 0.55 & 0.51 & 0.16\\

Ours & 853 & 19.42 & \textbf{\underline{5.72}} & 9.31 & \textbf{\underline{0.90}} & 0.14 & \textbf{\underline{0.51}} & \textbf{\underline{0.13}}\\
 
\bottomrule
\end{tabular}
}
\end{center}
\vspace{-0.2cm}
\caption{\label{tab:data statistic} Comparison of dataset statistics using various quality metrics.}
\vspace{-0.5cm}
\end{table*}

%%%%%%%%%%%%%%%%%%%%%%%

\begin{table*}[t]
\vspace{-0.2cm}
\begin{center}
\scalebox{0.84}{
\begin{tabular}{l||c|c|c||c|c|c|c||c|c|c|c}
\toprule
% \multirow{2}{*}{Type}
\textbf{Model} & \textbf{gleu $\uparrow$} & \textbf{meteor $\uparrow$} & \textbf{bs $\uparrow$} & \textbf{nov $\uparrow$} & \textbf{div $\uparrow$} & \textbf{dist-2 $\uparrow$} & \textbf{ent-2 $\uparrow$} & \textbf{arg $\uparrow$} & \textbf{c-arg $\uparrow$} & \textbf{cs $\uparrow$} & \textbf{tox $\downarrow$} \\
 
\midrule
\multicolumn{3}{l}{Vanilla} \\ 
\midrule
GPT3.5 & 0.011 & 0.087 & 0.834 & 0.877 & 0.658 & 0.565 & 10.352 & \textbf{\underline{0.280}} & 0.535 & 0.869 & 0.057 \\

GPT4o-mini & 0.008 & 0.070 & 0.829 & 0.891 & 0.578 & 0.496 & 10.424 & 0.241 & 0.627 & \textbf{0.891} & \textbf{0.042} \\

GPT4 & 0.009 & 0.075 & 0.830 & 0.888 & 0.624 & 0.521 & 10.845 &  0.162 & 0.560 & 0.793 & \textbf{\underline{0.039}} \\

Llama3 & 0.004 & 0.033 & 0.802 & 0.893 & 0.632 & 0.284 & \textbf{\underline{11.354}} & 0.101 & 0.482 & 0.773 & 0.053 \\

\midrule
\multicolumn{3}{l}{Strategy-Based Few-Shot} \\ 
\midrule

GPT3.5  & \textbf{\underline{0.019}} & \textbf{\underline{0.126}} & \textbf{\underline{0.838}} & \textbf{0.912} & \textbf{\underline{0.791}} & \textbf{\underline{0.728}} & 8.179 & 0.133 & 0.491 & 0.531 & 0.081 \\ 	

GPT4o-mini & 0.015 & 0.108  & \textbf{\underline{0.838}} & \textbf{0.912} & 0.756 & 0.663 & 8.493 & 0.171 & 0.568 & 0.716 & 0.055\\ 

GPT4 &   \textbf{0.017} & \textbf{0.113}  & \textbf{0.836} & \textbf{0.912} & \textbf{0.779} & \textbf{0.673} & 8.495 & 0.138 & 0.567 & 0.605 & 0.058\\ 

Llama3 & 0.004 & 0.034 & 0.791 & \textbf{\underline{0.922}} & 0.605 & 0.275 & 9.539 & 0.049 & 0.455 & 0.584 & 0.063\\ 

\midrule
\multicolumn{3}{l}{News Background}\\ 
\midrule
GPT3.5 & 0.009 & 0.074 & 0.831 & 0.886 & 0.665 & 0.520 & 10.450 & \textbf{0.245} & 0.556 & 0.882 & 0.053 \\ 

GPT4o-mini & 0.007 & 0.064 & 0.826 & 0.897 & 0.608 & 0.457 & 10.692 & 0.192 & \textbf{\underline{0.709}} & \textbf{\underline{0.936}} & \textbf{\underline{0.039}} \\ 

GPT4 & 0.007 & 0.066 & 0.828 & 0.881 & 0.678 & 0.475 & 11.152 & 0.141 & \textbf{0.631} & 0.762 & \textbf{\underline{0.039}} \\ 

Llama3 & 0.004 & 0.037 &  0.796 & 0.894 & 0.635 & 0.280 & \textbf{11.342} & 0.080 & 0.477 & 0.780 & 0.054 \\

\bottomrule
\end{tabular}
}
\end{center}
\vspace{-0.3cm}
\caption{\label{tab:results_all} Evaluation of responses in the settings of Vanilla, Strategy-Based Few-Shot, and adding News Background, regarding referential, diversity, and quality metrics.}
\vspace{-0.8cm}
\end{table*}

%%%%%%%%%%%%%%%%%%%%%%%

In our experiments, we use both closed-source and open-source large language models for counterspeech generation. Specifically, we employ close-source models gpt-3.5-turbo-0301, gpt4o-mini, and gpt4, and the open-source model Meta-Llama-3-8B-Instruct.

\vspace{-0.6cm}
\subsection{Evaluation}
\vspace{-0.2cm}
We used both the metrics and human judgment following \citep{saha2024crowdcounter} to evaluate our results from the perspectives of 
a) the correlation to the ground truth, 
b) the diversity and the novelty of the text, 
c) whether the generated reply has the quality of counterspeech and 
d) to what degree will it convince the offensive comment author. 
\subsubsection{Metrics}

\vspace{-0.6cm}
\stitle{\textit{Referential Metrics}} We used referential metrics to measure how similar the generated
counterspeech is to the ground truth in our dataset. We report two traditional metrics \textit{gleu} \cite{wu2016google}.
The similarity of the embedding between the generated text and the reference is also calculated by \textit{BERT Score} (\textit{bs}) \cite{zhang2019bertscore}.

\stitle{\textit{Diverity Metrics}} We employ two traditional diversity metrics \textit{dist-2} \cite{li2015diversity} and \textit{ent-2} \cite{baheti2018generating} to calculate its diversity. While \textit{dist-2} measures the proportion of distinct bigrams within the generated text and \textit{ent-2} calculates the text’s unpredictability and richness of word pairings, we also employ a semantic diversity (\textit{div}) and a semantic novelty (\textit{nov}) metric.

\stitle{\textit{Quality Metrics}} We evaluate key characteristics of counterspeech using the following metrics:

\textbf{\textit{Argument Quality (arg)}}. A good counter speech should be argumentative, it is measured by the confidence score of a \textit{roberta-base-uncased}\footnote{https://huggingface.co/chkla/roberta-argument} model fine-tuned by \citep{saha2024crowdcounter} on the argument dataset \cite{stab2018cross} on the generated counterspeech. 
\textbf{\textit{Counter-Argument Quality (c-arg)}}. This metric can measure the counter-argument degree of the abusive speech. We use the confidence score of a bert-base-uncased 
\footnote{https://huggingface.co/Hate-speech-CNERG/argument-quality-bert} model \cite{saha2024zero}, which can identify if the reply to an argument is a counter-argument or not. 
\textbf{\textit{Counterspeech Quality (cs)}}. This metric is applied to identify whether the text is counterspeech or not. It is useful because the generated counterspeech may differ from our ground truth in the strategy used, to avoid the low scores, as other metrics only calculate the correlation. The confidence score of the model based on bert-base-uncased \footnote{https://huggingface.co/Hate-speech-CNERG/counterspeech-quality-bert} is used \cite{saha2024zero}. 
\textbf{\textit{Toxicity (tox)}}. A counterspeech should be non-toxic itself. The used confidence score is based on the HateXplain model \cite{mathew2021hatexplain} and finetuned by \citep{saha2024crowdcounter} through the toxic data with two classes – toxic and non-toxic, and the toxic data contains both hate and offensive speech\footnote{https://huggingface.co/Hate-speech-CNERG/bert-base-uncased-hatexplain-rationale-two}.

\vspace{-0.6cm}
\subsubsection{Human Judgement}
In the process of Human Judgement, as we do in Section~\ref{subsec:huamn_annotation}, we also employed the evaluators from MTurk. The evaluators also help to annotate the strategies used in the text generated by vanilla LLM. 
Then we test the persuasiveness of the generated text by giving the evaluators the original comment, generated reply, together with the original reply to question them by \textit{If you are the sender of the comment, which reply can better convince you?}. 

\vspace{-0.5cm}

%%%%%%%%%%%%%%%%%%%%%%%%%%%%%
\begin{table}[t]
\vspace{-0.2cm}
\begin{center}
\scalebox{0.9}{
\begin{tabular}{l|c|c}
\toprule
\textbf{Strategy} & \textbf{Human} & \textbf{LLM}\\
\midrule

Affiliation & 3.04 & 3.05 \\
Counter question & 8.87 & 3.28 \\
Denouncing & 10.0 & 12.6 \\
Hostile language & 16.0 & 6.90 \\
Humor and sarcasm & 16.8 & 8.60 \\
Pointing out hypocrisy & 18.1  & 16.1 \\
Positive tone & 12.9 & 20.2 \\
Presenting facts & 10.0 & 13.7 \\
Warning of consequences & 3.46 & 7.92 \\
Others & 0.760 & 7.70 \\
\bottomrule
\end{tabular}}
\end{center}
\vspace{-0.2cm}
\caption{\label{tab:stat_counter}Proportion of different counterspeech strategies of our dataset (human written) and LLM generated.}
\vspace{-0.6cm}
\end{table}
% %%%%%%%%%%%%%%%%%%%%%%%%%%%%%

\section{Experimental Results}

\vspace{-0.3cm}
\subsection{Comparison Between Human and LLM}
%%%%%%%%%%%%%%
\vspace{-0.3cm}
For reference to the model-generated results, we show the statistics of our dataset and two baseline datasets in Table~\ref{tab:data statistic}. Two additional metrics \textit{Fleisch Kincaid} (\textit{fk})~\cite{flesch2007flesch} and \textit{Dale Chall} (\textit{dc})~\cite{dale1948formula} are used for verify the readability.

% \vspace{-0.2cm}
\begin{table*}[t]
\begin{center}
\vspace{-0.2cm}
\scalebox{0.82}{
\begin{tabular}{l||c|c|c||c|c|c|c||c|c|c|c}
\toprule
% \multirow{2}{*}{Type}
\textbf{Model} & \textbf{gleu $\uparrow$} & \textbf{meteor $\uparrow$} & \textbf{bs $\uparrow$} & \textbf{nov $\uparrow$} & \textbf{div $\uparrow$} & \textbf{dist-2 $\uparrow$} & \textbf{ent-2 $\uparrow$} & \textbf{arg $\uparrow$} & \textbf{c-arg $\uparrow$} & \textbf{cs $\uparrow$} & \textbf{tox $\downarrow$} \\
 
\midrule

GPT3.5 0-shot & 0.013 & 0.102 & 0.836 & 0.906 & 0.672 & 0.610 & \textbf{8.253} & \textbf{\underline{0.199}} & \textbf{\underline{0.530}} & \textbf{\underline{0.740}} & \textbf{\underline{0.070}} \\ 

GPT3.5 3-shot & \textbf{0.018} & 0.123 & \textbf{0.838} & \textbf{0.909} & \textbf{0.782} & \textbf{\underline{0.730}} & 8.199 & 0.155 & \textbf{0.511} & 0.542 & 0.088 \\ 

GPT3.5 5-shot & \textbf{\underline{0.019}} & \textbf{\underline{0.126}} & \textbf{0.838} & \textbf{\underline{0.912}} & \textbf{\underline{0.791}} & \textbf{0.728} & 8.179 & 0.133 & 0.491 & 0.531 & \textbf{0.081} \\ 	

GPT3.5 10-shot & \textbf{0.018} & \textbf{0.125} & \textbf{\underline{0.839}} & 0.903 & 0.778 & 0.720 & \textbf{\underline{8.590}} & \textbf{0.171} & 0.493 & \textbf{0.586} & 0.090 \\ 
 
\bottomrule
\end{tabular}
}
\end{center}
\vspace{-0.3cm}
\caption{\label{tab:res_fewshot}
Evaluation of responses with varying numbers of few-shot examples.}
\vspace{-1.1cm}
\end{table*}

The evaluations of vanilla responses are present in Table~\ref{tab:results_all}. We can observe that \textit{GPT} models perform better than \textit{llama} in the ability of basic text generation, with higher \textit{gleu}, \textit{meteor}, and \textit{dist-2}. \textit{llama} gets a higher \textit{ent-2} score, maybe because of its unpredictable phrase use, which cannot demonstrate its diversity.

However, in general, whether the model is open-source or closed-source, the results are similar but differ significantly compared with human-written replies. We notice that generated texts have lower diversity compared with our dataset, but they perform better with higher argumentativeness and lower toxicity.

The generated texts of \textit{GPT4} are evaluated in human evaluation, and only $4.84\%$ of the generated texts are evaluated as non-counterspeech. We list the strategy distribution in the Table~\ref{tab:stat_counter} and can observe that LLMs prefer to generate the counterspeech in a more patient and friendly way. We also calculate the length of generated replies, which is $46.1$, and is more than twice that of our dataset ($19.42$).
Significantly, the proportion of Positive Tone strategy has increased from $12.9\%$ to $20.2\%$, and Hostile Language has decreased from $16.0\%$ to $6.90\%$. Corresponding to our previous results, the generated replies lack diversity; they seldom use Counter Question to question the comment, and also, it is hard for them to apply Humor and Sarcasm.

In the persuasiveness test, $77.4\%$ of evaluators said they are more easily convinced by the LLM-generated reply if they are the senders of the harmful comments. However, although the friendly way is welcomed, the lack of diversity will affect its application in the real world.

\vspace{-0.5cm}
\subsection{Quality Improvements}
\vspace{-0.2cm}
\paragraph{\textbf{Results of Strategy-Based Few-Shot Prompting}}

Our results of the few-shot setting in Table~\ref{tab:results_all} show that adding several examples from human-written counterspeech and specifying the counterspeech strategy can effectively increase the novelty (\textit{nov}) and diversity (\textit{div} and \textit{dist-2}). Not surprisingly, the scores of the referential metrics (\textit{glue}, \textit{meteor}, and \textit{bs}) are all increasing. From results of different few-shot settings (Table~\ref{tab:res_fewshot}), we can observe 5-shot is an optimal setting for this task.

\vspace{-0.4cm}
\paragraph{\textbf{News Background Consideration}}
Table~\ref{tab:results_all} shows the evaluation of responses with news article information included in the prompt.
It is reasonable that the scores of similarity to the references decrease, as more knowledge of the LLM itself may be added. 
Compared with the vanilla setting, the \textit{nov} scores and \textit{div} scores have increased, which shows that the LLMs do not generate the counterspeech in the same template
At the same time, the scores of \textit{c-arg} and \textit{cs} have significantly improved, with \textit{tox} still kept at a low level, which indicates that the generated texts are effective counterspeech.

\vspace{-0.4cm}
\paragraph{\textbf{Results of  Different Strategies}}

We can observe that the performances differ in different strategies. For example, it is easy to generate similar counterspeech as the ground truth on the Hostile language strategy; the counterspeech of the Affiliation strategy has lower toxicity, while counterspeech with the Humor and sarcasm strategy has higher diversity; Counterspeech with the strategy of Presenting facts and Warning of consequences has higher argument and counter-argument quality.
The details of the results of the type-specific responses that specify a certain type of counterspeech strategy with news information are in Table~\ref{tab:evaluation_per_type} of Appendix~\ref{app:exp}.

\vspace{-0.5cm}
\subsection{Case Study}
\vspace{-0.2cm}
The counterspeech generated by vanilla models is almost the same, showing understanding and conviction in general. The selected examples are as follows:

\textit{\small I understand you're upset and we might disagree, but let's keep our conversation respectful and productive. We can learn a lot from each other when we communicate effectively.}

\textit{\small Let's remember that everyone deserves respect and understanding, regardless of gender. It's important to foster a culture of equality, where everyone's voice is valued and heard. Let's strive to appreciate and respect each other's differences.}

When the news background is added and examples are given, the model can target the event and combine the details of the comment to generate more powerful counterspeech. The selected examples are as follows:

\vspace{-0.3cm}
\paragraph{\textbf{Example 1}}
% \textbf{\textit{Example 1}}
\textbf{}

\textit{\small
\textbf{News Background}: Delta variant is spreading in California as COVID-19 battle enters an uncertain phase
}
\textit{\small
\textbf{Offensive comments}: There r idiots in California who don't/won't believe this Good! Do us all a favor. Minimalize covid again, or con't to believe covid never existed Do not waste a vaccine on yourself or you family If you do get the virus and if you die ... (yawn .. something..something)
}
\textit{\small
\textbf{Generated counterspeech}: Hey there, let's try to focus on the facts and science rather than resorting to name-calling. It's important for all of us to work together to stop the spread of COVID-19 and protect our communities. Let's stay informed and make responsible decisions for the well-being of everyone around us.
}
\vspace{-0.3cm}
\paragraph{\textbf{Example 2}}
% \textbf{\textit{Example 2}}
\textbf{}

\textit{\small
\textbf{News Background}: Former President Jimmy Carter lives in a \$167,000 house and shops at the Dollar General
}
\textit{\small
\textbf{Offensive comments}: Hmm Biden's family is owned by China and Ukraine millions have been wired to them fact! Trump Russia collusion was debunked. Democrats are so corrupt it is a threat to our national security! oh well
}
\textit{\small
\textbf{Generated counterspeech}: I appreciate your passion for politics, but let's remember that we are all Americans at the end of the day. Let's focus on finding common ground rather than spreading hate.
}

\vspace{-0.4cm}

\section{Conclusion}

\vspace{-0.3cm}
This work contributes a manually annotated dataset with $853$ pairs of offensive comments and counterspeech replies under the framework of media bias.
An analysis is conducted on the relationship between the harmful comments and the original news articles, also the counterspeech is explored in fine-grained strategies. At last, a benchmark of the new dataset is contributed to compare counterspeech generated by humans and large language models. Several methods, including few-shot learning and integration of news background information, are conducted to improve the diversity of the model-generated counterspeech.

\begin{credits}
\subsubsection{\ackname}
This work was partially supported by Hong Kong RGC GRF No. 14206324 and CUHK Knowledge Transfer Project Fund No. KPF23GWP20.
\end{credits}

% % \bibliography{mybibliography}
% \bibliography{custom}
\renewcommand{\bibname}{Reference}
\bibliographystyle{splncs04}
\bibliography{custom}

\appendix

\section{Experimental Results}
\label{app:exp}
The results of both adding news information and specifying a certain type of counterspeech strategy are in Table~\ref{tab:evaluation_per_type}.

%%%%%%%%%%%%%%%%%%%%%%%

%%%%%%%%%%%%%%%%%%%%%%%

% 
\begin{table*}[t]
\vspace{-0.1cm}
\begin{center}
\scalebox{0.68}{
\begin{tabular}{l|c||c|c|c||c|c|c|c||c|c|c|c}
\toprule
% \multirow{2}{*}{Type}
\textbf{Strategy Type} & \textbf{Model} & \textbf{gleu $\uparrow$} & \textbf{meteor $\uparrow$} & \textbf{bs $\uparrow$} & \textbf{nov $\uparrow$} & \textbf{div $\uparrow$} & \textbf{dist-2 $\uparrow$} & \textbf{ent-2 $\uparrow$} & \textbf{arg $\uparrow$} & \textbf{c-arg $\uparrow$} & \textbf{cs $\uparrow$} & \textbf{tox $\downarrow$} \\

\midrule

\multirow{4}{*}{Affiliation} & GPT3.5 & \textbf{0.026} & \textbf{0.141} & 0.842 & 0.917 & 0.833 & 0.763 & 7.851 & 0.134 & 0.615 & 0.621 & 0.049 \\
 & GPT4-mini & 0.019 & 0.122 & 0.839 & \textbf{\underline{0.936}} & 0.792 & 0.691 & 7.479 & 0.146 & 0.429 & 0.714 & \textbf{\underline{0.031}} \\
 & GPT4 & 0.015 & 0.099 & 0.832 & 0.924 & 0.792 & 0.675 & 7.435 & 0.199 & 0.512 & 0.429 & 0.048 \\
 & Llama3 & 0.00 & 0.04 & \textbf{\underline{0.94}} & 0.46 & 0.28 & \textbf{0.797} & 8.959 & 0.048 & 0.237 & 0.857 & \textbf{\underline{0.031}} \\
\hline

\multirow{4}{*}{Counter question} & GPT3.5 & 0.016 & 0.117 & 0.841 & 0.896 & 0.778 & 0.758 & 8.110 & 0.150 & 0.510 & 0.528 & 0.084 \\
 & GPT4-mini & 0.014 & 0.097 & 0.838 & 0.890 & 0.793 & 0.688 & 8.722 & 0.117 & 0.626 & 0.714 & 0.059 \\
 & GPT4 & 0.018 & 0.121 & 0.837 & 0.883 & 0.789 & 0.723 & 8.554 & 0.116 & 0.507 & 0.808 & 0.082 \\
 & Llama3 & 0.00 & 0.03 & 0.796 & 0.90 & 0.62 & 0.27 & 9.649 & 0.038 & 0.395 & 0.524 & 0.044 \\
\hline

\multirow{4}{*}{Denouncing} & GPT3.5 & 0.016 & 0.110 & 0.838 & 0.905 & 0.692 & 0.691 & 8.727 & 0.118 & 0.561 & 0.812 & 0.078 \\
 & GPT4-mini & 0.016 & 0.109 & 0.836 & 0.919 & 0.690 & 0.666 & 8.661 & 0.203 & 0.324 & 0.706 & 0.098 \\
 & GPT4 & 0.015 & 0.104 & 0.836 & 0.910 & 0.688 & 0.636 & 8.991 & 0.168 & 0.413 & 0.696 & 0.057 \\
 & Llama3 & 0.00 & 0.03 & 0.788 & \textbf{0.93} & 0.62 & 0.24 & 9.501 & 0.096 & 0.423 & 0.739 & \textbf{0.032} \\
\hline

\multirow{4}{*}{Hostile language} & GPT3.5 & \textbf{\underline{0.029}} & \textbf{\underline{0.159}} & 0.834 & 0.914 & 0.756 & \textbf{\underline{0.802}} & 8.825 & 0.098 & 0.620 & 0.249 & 0.127 \\
 & GPT4-mini & 0.024 & 0.139 & 0.835 & 0.915 & 0.729 & 0.732 & 9.333 & 0.106 & 0.595 & 0.519 & 0.058 \\
 & GPT4 & 0.024 & 0.133 & 0.833 & 0.917 & 0.731 & 0.725 & 9.430 & 0.085 & 0.573 & 0.378 & 0.123 \\
 & Llama3 & 0.00 & 0.03 & \textbf{0.91} & 0.785 & 0.58 & 0.25 & 10.270 & 0.035 & 0.506 & 0.567 & 0.088 \\
\hline

\multirow{4}{*}{Humor and sarcasm} & GPT3.5 & 0.018 & 0.134 & 0.831 & 0.905 & \textbf{\underline{0.857}} & 0.783 & 9.212 & 0.081 & 0.521 & 0.508 & 0.157 \\
 & GPT4-mini & 0.015 & 0.112 & 0.828 & 0.902 & 0.845 & 0.721 & 9.692 & 0.148 & 0.659 & 0.640 & 0.078 \\
 & GPT4 & 0.014 & 0.109 & 0.827 & 0.905 & \textbf{0.846} & 0.738 & 9.667 & 0.064 & 0.597 & 0.513 & 0.087 \\
 & Llama3 & 0.01 & 0.03 & 0.783 & 0.90 & 0.63 & 0.28 & 10.104 & 0.042 & 0.391 & 0.610 & 0.076 \\

\hline
\multirow{4}{*}{Pointing out hypocrisy} & GPT3.5 & 0.016 & 0.118 & 0.844 & 0.889 & 0.809 & 0.677 & 9.852 & 0.212 & 0.506 & 0.655 & 0.084  \\
 & GPT4-mini & 0.016 & 0.119 & 0.845 & 0.884 & 0.795 & 0.648 & 9.873 & 0.207 & 0.519 & 0.671 & 0.063 \\
 & GPT4 & 0.015 & 0.101 & 0.841 & 0.895 & 0.808 & 0.620 & 10.050 & 0.208 & 0.580 & 0.740 & 0.055 \\
 & Llama3 & 0.00 & 0.03 & 0.797 & 0.91 & 0.56 & 0.25 & \textbf{\underline{10.659}} & 0.043 & 0.266 & 0.719 & 0.091 \\

\hline
\multirow{4}{*}{Positive tone} & GPT3.5 & 0.013 & 0.097 & 0.836 & 0.916 & 0.673 & 0.621 & 8.731 & 0.117 & 0.600 & 0.664 & 0.045 \\
 & GPT4-mini & 0.011 & 0.090 & 0.837 & 0.912 & 0.680 & 0.588 & 9.193 & 0.118 & 0.601 & \textbf{\underline{0.900}} & 0.034 \\
 & GPT4 & 0.012 & 0.094 & 0.837 & 0.911 & 0.680 & 0.580 & 9.103 & 0.079 & 0.585 & 0.683 & 0.038 \\
 & Llama3 & 0.00 & 0.03 & 0.798 & 0.92 & 0.59 & 0.28 & \textbf{10.368} & 0.058 & 0.619 & 0.567 & 0.062 \\

\hline
\multirow{4}{*}{Presenting facts} & GPT3.5 & 0.014 & 0.105 & 0.835 & 0.902 & 0.831 & 0.637 & 9.057 & \textbf{0.317} & \textbf{0.697} & 0.698 & 0.068 \\
 & GPT4-mini & 0.009 & 0.082 & 0.832 & 0.910 & 0.819 & 0.574 & 9.426 & \textbf{\underline{0.325}} & 0.492 & 0.796 & 0.065 \\
 & GPT4 & 0.008 & 0.074 & 0.835 & 0.905 & 0.814 & 0.518 & 9.732 & 0.158 & 0.526 & 0.785 & 0.039 \\
 & Llama3 & 0.00 & 0.03 & 0.793 & 0.92 & 0.62 & 0.29 & 9.063 & 0.065 & 0.606 & 0.474 & 0.047 \\

\hline
\multirow{4}{*}{Warning of consequences} & GPT3.5 & 0.021 & 0.129 & 0.851 & 0.927 & 0.754 & 0.712 & 7.375 & 0.241 & 0.410 & 0.622 & 0.077 \\
 & GPT4-mini & 0.014 & 0.109 & 0.847 & 0.916 & 0.720 & 0.606 & 7.976 & 0.227 & 0.623 & 0.737 & 0.072 \\
 & GPT4 & 0.014 & 0.101 & 0.843 & 0.925 & 0.750 & 0.637 & 7.342 & 0.227 & 0.251 & \textbf{0.875} & 0.037 \\
 & Llama3 & 0.00 & 0.03 & 0.787 & \textbf{0.93} & 0.59 & 0.25 & 9.296 & 0.024 & \textbf{\underline{0.750}} & 0.868 & 0.034 \\
 
\bottomrule
\end{tabular}
}
\end{center}
% \vspace{-0.2cm}
\caption{\label{tab:evaluation_per_type}Evaluation of type-specific responses with news article information regarding referential, diversity, and quality metrics. 
We report the type-specific scores.
}
\vspace{-0.4cm}
\end{table*}

%%%%%%%%%%%%%%%%%%%






% LNCS Homepage, \url{http://www.springer.com/lncs}, last accessed 2023/10/25
% \end{thebibliography}

% \begin{credits}
% \subsubsection{\discintname}
% The authors have no competing interests to declare that are relevant to the content of this article.
% \end{credits}

\end{document}